\documentclass{article}
\usepackage{ijcai22}

\usepackage{times}
\usepackage{soul}
\usepackage{url}
\usepackage[hidelinks]{hyperref}
\usepackage[utf8]{inputenc}
\usepackage[small]{caption}
\usepackage{graphicx}
\usepackage{amsmath}
\usepackage{amsthm}
\usepackage{booktabs}
\usepackage{algorithm}
\usepackage{algorithmic}
\usepackage{amsfonts}

\usepackage{subcaption}

\newcommand{\mcD}{\mathcal{D}}
\newcommand{\calD}{\mathcal{D}}

\title{Robust One Round Federated Learning with Predictive Space Bayesian Inference}

\author{
Mohsin Hasan$^{1,2}$
\and
Zehao Zhang$^{1,2}$\and
Kaiyang Guo$^{3}$\and
Mahdi Karami$^3$ \and \\
Guojun Zhang$^3$ \and
Xi Chen$^3$ \and
Pascal Poupart$^{1,2}$ \\
\affiliations
$^1$University of Waterloo, Cheriton School of Computer Science \\
$^2$Vector Institute\\
$^3$Huawei Noah's Ark Lab\\
\emails
\{mohsin.hasan, z783zhang, ppoupart\}@uwaterloo.ca, \\
\{guo.kaiyang, mahdi.karami1, guojun.zhang,
xi.chen4\}@huawei.com
}

\begin{document}

\maketitle

\begin{abstract}
    Making predictions robust is an important challenge. A separate challenge in federated learning (FL) is to reduce the number of communication rounds, particularly since doing so reduces performance in heterogeneous data settings. To tackle both issues, we take a Bayesian perspective on the problem of learning a global model. We show how the global predictive posterior can be approximated using client predictive posteriors. This is unlike other works which aggregate the local model space posteriors into the global model space posterior, and are susceptible to high approximation errors due to the posterior's high dimensional multimodal nature. In contrast, our method performs the aggregation on the predictive posteriors, which are typically easier to approximate owing to the low-dimensionality of the output space. We present an algorithm based on this idea, which performs MCMC sampling at each client to obtain an estimate of the local posterior, and then aggregates these in one round to obtain a global ensemble model. Through empirical evaluation on several classification and regression tasks, we show that despite using one round of communication, the method is competitive with other FL techniques, and outperforms them on heterogeneous settings. The code is publicly available at \url{https://github.com/hasanmohsin/FedPredSpace_1Round}.   
\end{abstract}

\section{Introduction}

Federated learning (FL) is a machine learning paradigm in which the goal is to train a model using decentralized data stored on client devices, and with the constraint that client data is kept private~\cite{mcmahan17a}. This paradigm is applicable to scenarios such as training a global model using data stored on several mobile devices. In most FL techniques, multiple rounds of communications are performed. In each round, the server sends the current global model to the clients, which update it by training with their local data. The clients then return the updated models to the server which aggregates them into a revised global model.  Two challenges arise:  i) How to reduce the number of communication rounds since each round has a heavy communication cost and requires a certain degree of synchronization? ii) How to ensure {\bf robustness} in the predictions made by the aggregated global model? Since the data and resources across clients may be heterogeneous, and since the local data distributions may change over time, it is useful for the global model to output a calibrated uncertainty estimate on its predictions. In this sense, a robust model is particularly important in the case of FL. Furthermore, this heterogeneity can negatively impact the aggregation of local models due to divergence between model parameters~\cite{zhao2018federated}. A simple strategy to mitigate divergence between local models is to increase the frequency of communication rounds, but this increases costs.  In this paper, we tackle the following question: How can we design a robust FL technique that performs a single round of communication?

We take a Bayesian approach to FL. As shown by \cite{ep_mcmc}, it is possible in principle to do distributed Bayesian learning in one round of communication. In Bayesian FL, clients estimate local posteriors over models based on their local data.  Then the server can aggregate the local posteriors into an exact global posterior in one round simply by multiplying the local posteriors (with a corrective factor depending on the prior).  Note that data and client heterogeneity does not impact this computation.  Furthermore, the posterior can be used to provide confidence estimates of the predictions which enables decision makers to decide when to trust a prediction.   The downside of this technique is that aggregating the local posteriors in this way requires approximating their densities. This approximation is subject to the curse of dimensionality in the sense that the quality of the approximation tends to decrease with the number of dimensions of the posterior. Here the dimensionality of a local posterior is determined by the number of parameters of the underlying model.

We propose a new Bayesian FL technique that operates over the predictive space instead of the model space.  Ultimately, the goal is to make predictions. In Bayesian learning, we typically estimate a posterior distribution over models, which is then integrated out to estimate a posterior distribution over predictions.  While the posterior distribution over models has high dimensionality (i.e., the number of dimensions corresponds to the number of model parameters), the posterior over predictions has much lower dimensionality since the dimensionality corresponds to the number of outputs which is often 1 or quite small. Hence, we design a Bayesian FL technique that estimates and then aggregates local predictive posteriors into a global predictive posterior. More precisely, we use a Markov Chain Monte Carlo (MCMC) sampling technique to estimate the local predictive posteriors.  Fewer samples are needed to estimate predictive posteriors than model posteriors due to the reduced dimensionality. Then we aggregate the local predictive posteriors by forming a weighted ensemble that takes into account the degree of confidence of each local prediction. 
The contributions of the paper can be summarized as follows:
\begin{itemize}
    \item A new Bayesian FL technique that operates directly in the space of predictive posteriors. This technique needs only one round of communication and scales better to larger models since it avoids the curse of dimensionality of model posteriors.
    \item Empirical evaluation of the new Bayesian FL technique on classification and regression tasks.  Despite the single round of communication, our technique is competitive with other FL techniques and outperforms them in heterogeneous settings. 
\end{itemize}

\section{Background and Related Work}

\paragraph{Federated Learning.} In federated learning (FL), the data is distributed across several clients. Let $\mathcal{D}=\mathcal{D}_1 \cup ... \cup \mathcal{D}_n$ where $\mathcal{D}_i = \{(x_1,y_1),...,(x_{k_i},y_{k_i})\}$ is the dataset of size $k_i$ at client $i$. The goal is to learn a predictive model without any data leaving their client to preserve privacy. Let $m:W,X \rightarrow \Delta_Y$ be a probabilistic predictive model parameterized by $w\in W$ that maps an input $x \in X$ to a distribution in the space of distributions $\Delta_Y$ (i.e., $m_w(x) = p(y|x,w)$).  To avoid sharing the data, a common approach consists of having each client learn a local model $m_{w_i}$ that is shared with a trusted server that aggregates the local models into a global model $m_{\bar{w}}$.  FedAveraging~\cite{mcmahan17a} aggregates local models by taking the average of their parameters (i.e., $\bar{w} = \sum_i w_i k_i/(\sum_i k_i)$).  

In practice, the datasets are often heterogeneous, which means that their content may be sampled from different distributions (i.e., for $(x,y)_i \in \mcD_i$ and $(x,y)_j \in \mcD_j$ we have $(x,y)_i \sim P_i(x,y)$ and $(x,y)_j \sim P_j(x,y)$ but $P_i \neq P_j$) and the amount of data at each client may differ (i.e.,  $\exists i\neq j$ such that $k_i \neq k_j$).  To deal with heterogeneity and avoid client divergence, FedAveraging~\cite{mcmahan17a} and many other variants~\cite{li2020federated,wang2020tackling,wang2019federated,li2020fedbn,mohri2019agnostic} perform frequent rounds of model updating and averaging where in each round the clients update their local models based on a few steps of gradient descent or more generally some form of partial training.  Unfortunately, this can be quite costly due to the increased communication and the need for synchronization at each round.

\paragraph{Bayesian Learning.} Bayesian techniques allow for the training of a model which provides robust uncertainty estimates. It does so by constructing a model which can make predictions using the \textbf{predictive posterior distribution}, denoted: $p(y|x, \mcD)$. Typically this is done by setting a ``model space prior" $p(w)$, and then obtaining approximate samples from the global model space posterior $\{w_j\} \sim p(w|\mcD) \propto p(w)p(\mcD| w)$. These samples are then used to estimate the integral:

\begin{align}
    p(y|x,\mcD) &= \int p(y|x, w) p(w|\mcD) dw \nonumber \\
                &\approx \sum_j p(y|x, w_i) 
                \label{eq:pred}
\end{align}

For a well chosen prior, this method can yield predictions with accurate uncertainty estimates.

\paragraph{Bayesian techniques in FL.} Existing Bayesian FL techniques focus on approximating the global model space posterior $p(w|\mcD)$ from the local model space posteriors $p(w|\mcD_i)$ ~\cite{ep_mcmc,al-shedivat2021federated}.

``Embarrasingly Parallel MCMC"~\cite{ep_mcmc} does so by drawing MCMC samples from each local posterior (with a corrective factor from the prior), and then estimating the local densities either as Gaussians or with a Kernel density estimator. These local densities are then aggregated via multiplication (again with a prior corrective factor) to obtain an approximation for the global model space posterior. This global density is then sampled to obtain the desired posterior samples to which Equation \eqref{eq:pred} may be applied for inference. It is worth noting that the original work wasn't designed for use with neural networks, and the memory costs associated with the method make it intractable for this setting. For instance when approximating the local posteriors as Gaussians and aggregating them, a computational cost of $O(d^3)$ is required for inverting the covariance matrices, where $d$ is the number of neural network parameters. This method is notable however for operating with only a single communication round.

``Federated Posterior Averaging"~\cite{al-shedivat2021federated} is similar to the above technique, except that it approximates the local posteriors as Gaussians, and devises a more efficient, iterative algorithm for aggregating the local posteriors (with cost linear in the number of network parameters). This method also operates in multiple rounds of communication. 

The main issue with both these techniques is that they require some approximation of the global model space posterior (e.g., in the form of a Gaussian), which can often be inaccurate when the number of model parameters is large. Such approximations are especially poor for neural network models, where the model space posterior is known to be multimodal~\cite{pourzanjani2017improving}. 

\section{Method}

We take the perspective of learning a global Bayesian model on a dataset composed of $n$ shards: $\mcD = \mcD_1 \cup ... \cup \mcD_n$, each stored on an individual client. We assume we are in the supervised setting so that each datapoint is an input and output pair $(x,y)$. Ultimately we'd like to construct a model which can make predictions using the predictive posterior $p(y|x, \mcD)$.

Towards this end, we assume the data shards are independent, i.e., $p(\mcD) = \prod_i p(\mcD_i)$ and conditionally independent given a datapoint $(x,y)$: $p(\mcD|y,x) = \prod_i p(\mcD_i | y,x)$. Note that each distribution need not be identical. Then, the global predictive posterior can be written:

\begin{align}
    p(y | x, \calD) &= p(y | x, \calD_1, ..., \calD_n) \nonumber\\
    &= p(\calD_1, ... , \calD_n | y, x) \frac{p(y|x)}{p(\calD)} \nonumber\\
    &= \frac{p(y|x)}{p(\calD)}\prod_i p(\calD_i | y, x) \nonumber\\
    &= \frac{p(y|x)}{p(\calD)} \prod_i p(y | x, \calD_i) \frac{p(\calD_i)}{p(y|x)} \nonumber\\
    &= \frac{1}{p(y|x)^{n-1}} \prod_i p(y | x,\calD_i) \label{eq:prod_rule}
\end{align}

Where $p(y|x)$ is the ``prior predictive distribution", and is determined by the chosen model space prior $p(w)$.

Assuming that each client is able to provide some approximation to its local predictive posterior $p(y|x,\mcD_i)$, Equation \eqref{eq:prod_rule} can be interpreted as an aggregation technique. To proceed further we must make some assumptions on the form of the predictions.

\subsection{Aggregation for Regression}
\label{sec:agg_regr}

For a regression task, suppose $y \in  \mathbb{R}^d$. In this case, we can approximate the local predictive posteriors as (multivariate) Gaussians: $p(y|x, \mcD_i) = \mathcal{N}(\mu_i, \Sigma_i)$ (with $\mu_i,~\Sigma_i$ depending on $x$). We similarly approximate the prior predictive distribution $p(y|x) = \mathcal{N}(\mu_p, \Sigma_p)$. 

Since the aggregation formula \eqref{eq:prod_rule} multiplies or divides these densities, the global predictive posterior will also be a Gaussian with some mean $\mu_g$ and covariance $\Sigma_g$:

\begin{align}
    \Sigma_g &= \Big(\sum_i \Sigma^{-1}_i - (n-1)\Sigma^{-1}_p \Big)^{-1}
    \label{eq:var_regr}
\end{align}

\begin{equation}
   \mu_g = \Sigma_g (\sum_i \Sigma^{-1}_i \mu_i + (n-1)\Sigma^{-1}_p\mu_p)
   \label{eq:mean_regr}
\end{equation}

The required means and covariances in these formulas may be estimated given samples from each predictive distribution, which in turn may be obtained from samples from the model space posterior $p(w|\mcD_i)$ using any MCMC method.

Note that this aggregation formula has an intuitive interpretation. Suppose we are in the one-dimensional setting, where $\Sigma_i = \sigma^2_i$ is the variance of the output from the client $i$. Further suppose that we have selected a prior with high uncertainty $\Sigma_p = \sigma^2_p$, and mean $\mu_p = 0$ (which are reasonable settings for an uninformative prior) so that we can ignore these terms in the formula. Then the aggregation formulas in \eqref{eq:var_regr} and \eqref{eq:mean_regr} become the weighted sum:

\begin{align}
 \mu_g &= \sum_i r_i \mu_i \nonumber \\
 r_i &= \frac{(\sigma^2)^{-1}_i}{\sum_i (\sigma^{2})^{-1}_i}
\end{align}

The weight $r_i$ characterizes the uncertainty client $i$ has in its prediction at input $x$. A client with high uncertainty would have a correspondingly low weight, and therefore less influence on the overall (mean) prediction. This is a helpful feature in settings with heterogeneous data. In these settings, a client dataset $\mcD_i$ may not contain any data resembling, or close to some query input $x$, and we wouldn't like the global prediction at $x$ to be influenced by such clients.

\paragraph{Justification for Gaussian Approximation.}

\begin{table*}[t]
    \centering
    \begin{tabular}{lccccc}
        \toprule
        Method  & MNIST & Fashion MNIST & EMNIST & CIFAR10 & CIFAR100 \\
        \midrule
        FedAvg   &    \textbf{97.68} $\pm$ 1.05       & 85.93 $\pm$ 2.14   & \textbf{87.53} $\pm$ 1.22 &  \textbf{80.70} $\pm$ 0.20 & \textbf{42.52} $\pm$ 0.44  \\
        FedPA & 96.67 $\pm$ 0.28          & 83.67  $\pm$ 2.28    & 85.38 $\pm$ 0.44  & 61.35 $\pm$ 0.28   &  35.36 $\pm$ 1.78   \\ 
        \midrule
        FedAvg & 94.48 $\pm$ 2.11 & 84.12 $\pm$ 3.35 & 84.23 $\pm$ 0.48 & 73.55 $\pm$ 0.10 & 27.41 $\pm$ 0.50\\
        EP MCMC & 95.34 $\pm$ 0.14 & 83.30 $\pm$ 1.53  & 83.79 $\pm$ 0.87  &  73.93 $\pm$ 0.68 &    35.66 $\pm$ 1.32  \\
        PredictiveBayes (\textbf{ours}) & \textit{97.28} $\pm$ 0.19 & \textit{\textbf{86.38}} $\pm$ 0.49    &  \textit{86.51} $\pm$ 1.22  & \textit{76.83} $\pm$ 1.63    & \textit{42.06} $\pm$ 1.60 \\
        \bottomrule
    \end{tabular}
    \caption{Test accuracies on classification datasets, based on 3 seeds. \textbf{Higher is better}. Multi-round methods are written above the line, while methods run for a single round of communication are written below. The best technique overall is bolded, while the best technique over 1 round is italicized.}
    \label{tab:class_acc}
\end{table*}

\begin{table*}[t]
    \centering
    \begin{tabular}{lccccc}
        \toprule
        Method  & Air Quality & Bike & Wine Quality & Real Estate & Forest Fire \\
        \midrule
        FedAvg   & 1.070 $\pm$ 0.022          & 0.0814 $\pm$ 0.0150  & 0.1180 $\pm$ 0.0030  & 0.0196 $\pm$ 0.0032 &  0.1142 $\pm$ 0.0080   \\
        FedPA &  1.567 $\pm$ 0.237        & 0.6070 $\pm$ 0.0133   & 0.2180 $\pm$ 0.0224  & 0.0303 $\pm$ 0.0054  & \textbf{0.0892} $\pm$ 0.0010    \\ 
        \midrule
        FedAvg & 1.050 $\pm$ 0.016 & 0.0826 $\pm$ 0.0200 & 0.1190 $\pm$ 0.0025 & \textit{\textbf{0.0185}} $\pm$ 0.0040 & 0.1050 $\pm$ 0.0080 \\
        EP MCMC &  \textit{\textbf{0.903}} $\pm$ 0.020         & 0.0667 $\pm$ 0.0150  &  0.1174 $\pm$ 0.0050 & 0.0190 $\pm$ 0.0030  &    \textit{0.0954} $\pm$ 0.0010  \\
        PredictiveBayes (\textbf{ours}) & 0.909 $\pm$ 0.002 &  \textit{\textbf{0.0587}} $\pm$ 0.0120  & \textit{\textbf{0.1149}} $\pm$ 0.0040    &    0.0188 $\pm$ 0.0030 & 0.0968 $\pm$ 0.0070 \\
        \bottomrule
    \end{tabular}
    \caption{Test mean square error on regression datasets, based on 3 seeds. \textbf{Lower is better}. Multi-round methods are written above the line, while methods run for a single round of communication are written below. The best technique overall is bolded, while the best technique over 1 round is italicized.}
    \label{tab:regr_acc}
\end{table*}

The Gaussian approximation may seem like a severe approximation, but we argue that it is reasonable for predictive posteriors.  Generally speaking, approximating a distribution by a Gaussian is reasonable when the distribution is unimodal and we assume a loss based on the squared distance to a unique target value.  Recall that the predictive posterior $p(y|x,\mathcal{D}_i)$ is a distribution over output values $y$.  In supervised regression, we typically assume that there is a single target value $y^*$ and we often seek to minimize the squared error $(y-y^*)^2$. Similarly, under suitable conditions, Bayesian consistency~\cite{nogales2022consistency} ensures that the expectation of the predictive posterior will converge to the target value $y^*$ in probability (i.e., $E_{p(y|x,\mathcal{D}_i)}[y] \rightarrow y^*$). 
In the case of a Gaussian predictive posterior, the probability that a prediction $y$ is correct is proportional to the exponential of the squared distance to the expectation (i.e.,  $p(y|x,\mathcal{D}_i) \propto \exp({(y-E_{p(y|x,\mathcal{D})}[y])^2})$.  Hence, the assumption of a Gaussian predictive posterior is in line with the assumption of a unique target $y^*$ and the minimum squared error in supervised regression.

In contrast, assuming a Gaussian posterior $p(w|\mathcal{D}_i)$ in the model space would not be reasonable since there are typically many equivalent models (due to symmetries) that can generate the same data $\mathcal{D}_i$.  For instance, if we consider the space of neural networks with fully connected layers, it is well known that hidden nodes can be interchanged to obtain symmetrically equivalent models~\cite{pourzanjani2017improving}. Hence the model posterior $p(w|\mathcal{D}_i)$ is typically multimodal and far from Gaussian. 

\subsection{Aggregation for Classification}
\label{sec:agg_class}

For classification: $y$ is discrete. The product in (\ref{eq:prod_rule}) can be computed directly, i.e., for each value of $y \in \{l_1, ..., l_K\}$ (where $l_i$ is the $i$th class label):

\begin{equation}
    p(y = l_j|x,\mcD) = \frac{1}{p(y=l_j|x)^{n-1}} \prod_i p(y = l_j | x, \mcD_i)
    \label{eq:class_pred}
\end{equation}

We can interpret this in terms of uncertainties. First, rewriting the formula as:

\begin{flalign*}
 p(y|x,\mcD) =  p(y|x) \prod_i \frac{p(y|x,\mcD_i)}{p(y|x)}
\end{flalign*}

Each client contributes a factor of $\frac{p(y|x,\mcD_i)}{p(y|x)}$ (the quotient of the posterior and prior in predictive space). If client $i$ doesn't learn much, and has little data (has high uncertainty), its local posterior will be closer to the prior. Thus the factor $\frac{p(y|x,\mcD_i)}{p(y|x)} \approx 1$ for each $y$. This means the factor does not contribute much to the overall prediction.\\

\subsection{Algorithm}

Given the aggregation formulas above, the overall algorithm (called ``predictive space Bayes", or PredictiveBayes) consists of the steps:

\begin{enumerate}
    \item At each client, use MCMC sampling to generate samples according to the local posteriors $\{w_{j}\} \sim p(w|\mcD_i)$ 
    
    \item Have each client communicate the samples to each other
    
    \item At prediction time, use each set of samples to produce predictions according to $p(y|x,\mcD_i)$ (using Equation \eqref{eq:pred})
    
    \item Using Equation \eqref{eq:mean_regr} (for regression tasks) or \eqref{eq:class_pred} (for classification tasks), aggregate the individual predictions into the global predictive posterior $p(y|x,\mcD)$.
\end{enumerate}

Since the aggregation is done at the predictive space level, the algorithm essentially builds an ensemble of models to predict according to the global posterior. 

Note that the algorithm works with a single round of communication. This is because the sampling from the local posteriors can be done individually by each client, and the aggregation step computes $p(y|x,\mcD)$ in one step. This feature of the method alleviates many common problems faced by other FL techniques, such as synchronization issues, and the heavy cost of communication that multiple rounds bring with them.

\section{Experiments}

We verify the effectiveness of our method by training it on multiple regression and classification datasets, and comparing its performance to a selection of baseline algorithms. All tests were run across 5 clients.

\subsection{Experimental Setup}

\begin{figure*}
\centering
\begin{subfigure}{0.333\textwidth}
  \centering
  \includegraphics[width=\linewidth]{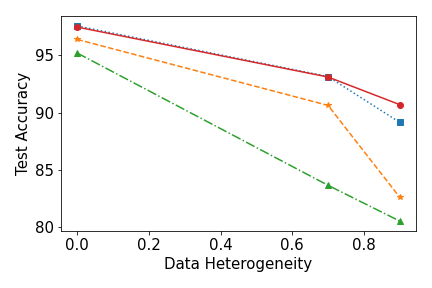}
  \caption{MNIST}
\end{subfigure}%
\begin{subfigure}{0.333\textwidth}
  \centering
  \includegraphics[width=\linewidth]{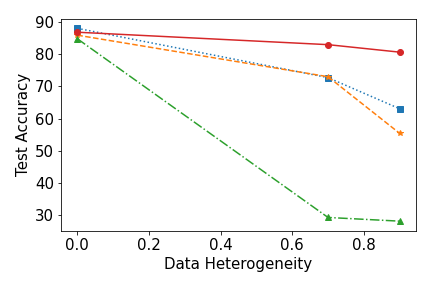}
  \caption{Fashion MNIST}
\end{subfigure}%
\begin{subfigure}{0.333\textwidth}
  \centering
  \includegraphics[width=\linewidth]{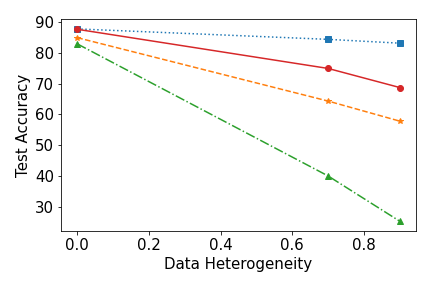}
  \caption{EMNIST}
\end{subfigure}\par\medskip
\begin{subfigure}{0.333\textwidth}
  \centering
  \includegraphics[width=\linewidth]{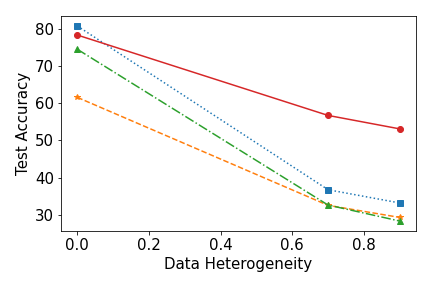}
  \caption{CIFAR10}
\end{subfigure}
\begin{subfigure}{0.333\textwidth}
  \centering
  \includegraphics[width=\linewidth]{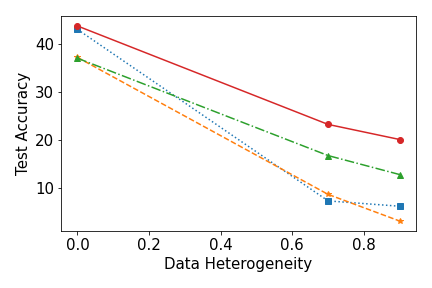}
  \caption{CIFAR100}
\end{subfigure}
\begin{subfigure}{0.3\textwidth}
  \centering
  \includegraphics[width=0.8\linewidth]{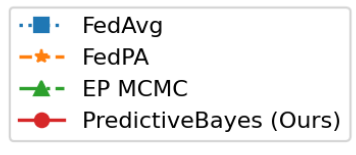}
  \caption{Legend}
\end{subfigure}
\caption{Test accuracies on the classification datasets with increasing data heterogeneity (tested with parameter settings of 0.0, 0.7 and 0.9)}
\label{fig:noniid_class}
\end{figure*}

Datasets were split into a global train and test set. The train set was further divided among clients. All reported results correspond to the test set performance.

\paragraph{Classification Datasets.} To evaluate the method for classification, the following datasets were chosen: EMNIST (with 62 classes), MNIST, Fashion MNIST, CIFAR10, and CIFAR100. The way in which each of these datasets are distributed among clients is controlled by a ``heterogeneity parameter". For homogeneous data (or a parameter value of 0), the data is split uniformly to each client, whereas in the maximally heterogeneous (or fully non-iid) setting (with a parameter value of 1.0), the data is sorted by class before being split among clients. This means that each client observes data from different classes with little overlap. In this heterogeneous setting, the aggregation technique plays a key role in constructing a good global model, since each local model performs poorly given the imbalanced training data. Values of the parameter between 0 and 1 mix data from either setting, by replacing the corresponding fraction of the iid data shard with the fully non-iid data for each client.

\paragraph{Regression Datasets.} The regression datasets used for evaluation include: the ``wine quality", ``air quality", ``forest fire", ``real estate", and ``bike rental" datasets from the UCI repository~\cite{uci}. These datasets were split among clients uniformly.

\paragraph{Models.} For the regression datasets, MNIST, Fashion MNIST and EMNIST a two layer fully connected network was used, with 100 hidden units. For the CIFAR10 and CIFAR100 datasets a Convolutional Neural network was used with 3 convolution layers, each followed by ``Max Pooling" layers, with a single fully connected layer at the end. For all networks, the ReLU activation function was used.

\paragraph{Baselines.} The Federated techniques compared to include: Federated Averaging (FedAvg)~\cite{mcmahan17a}, Federated Posterior Averaging (FedPA)~\cite{al-shedivat2021federated}, and Embarrassingly Parallel MCMC (EP MCMC)~\cite{ep_mcmc}. In the case of FedAvg, SGD with momentum was used for local optimization. For the rest of the methods, including our own, since they require MCMC sampling, cyclic stochastic gradient Hamiltonian Monte-Carlo (cSGHMC)~\cite{zhang2020csgmcmc} was used. For EP MCMC, the algorithm was computationally intractable for neural network models due to requiring the calculation of the inverse of a covariance matrix over parameters. Thus, for these experiments, a diagonal covariance matrix was assumed (which corresponds to the assumption that the local posteriors are approximated by an axis-aligned Gaussian). FedAvg and FedPA are methods meant to run for multiple rounds, whereas EP MCMC and our method are meant to run in 1 round. To better compare the influence of communication rounds, we ran FedAvg for both multiple rounds and with only 1 round of communication. 

\paragraph{Training Details.} For MNIST, Fashion MNIST and EMNIST, the training was run for 25 epochs per client overall (split into 5 rounds for multi-round methods, while only run in a single round for EP MCMC and our method). For CIFAR10 and CIFAR100, training was run for 50 epochs per client (split into 10 rounds for multi-round methods), and for the regression datasets, training was run for 25 epochs per client (again split into 5 rounds for multi-round methods). The methods using sampling (FedPA, EP MCMC and PredictiveBayes) used 10 samples for all experiments.

\subsection{Results}

\subsubsection{Classification Results}

The results on the classification setting are recorded in Table \ref{tab:class_acc}. For all datasets, we can identify that among the single round methods, our method performs best. Compared to the multi-round methods, our method is second-best (behind FedAvg), with an accuracy difference of $4\%$ for CIFAR10 and a difference of under $1\%$ for all other datasets. These results suggests that despite training for only a single round, our method is competitive with more widely used multi-round techniques.

The discussion in Section \ref{sec:agg_class} suggests that our method would provide some advantage in the heterogeneous setting. We verify this intuition by comparing the performance of the different algorithms as the level of data heterogeneity is increased. The results are plotted in Figure \ref{fig:noniid_class}. For these results, FedAvg and FedPA were run for multiple rounds (5 for the MNIST-like datasets, and 10 for the CIFAR datasets), while EP MCMC and PredictiveBayes were run for a single round. 

From these results, we can see that in all datasets except for EMNIST, our method outperforms the other techniques as the heterogeneity increases. This is particularly important since in realistic federated learning scenarios, data is often heterogeneous. 

\subsubsection{Regression Results}

The results for the regression setting are recorded in Table \ref{tab:regr_acc}. We can again observe that for most datasets, our method either performs best, or has a small relative performance gap behind the best method (either EP MCMC or FedAvg).

\section{Limitations and Future Work}

\paragraph{Privacy.} Our algorithm makes use of MCMC techniques to sample from a Bayesian posterior. In the literature, there are differentially private variants for MCMC techniques ~\cite{dp_mcmc,dp_hmc} which may be substituted into the presented method. Further study is required to understand the performance trade-offs (if any), and to obtain a better understanding of our algorithm's influence on privacy. 

\paragraph{Scalability.} The presented algorithm (PredictiveBayes) works by creating an ensemble model with members being samples from individual client posteriors. This means that the computational cost of inference scales linearly with the number of clients. Since most practical FL scenarios can involve thousands of clients~\cite{leaf}, this computational cost can become a bottleneck. Future improvements to this method can focus on approximating the ensemble with a fixed size model, so that the inference cost remains fixed with the number of clients.

\section{Conclusion}

In this work, we presented a Bayesian technique for federated learning which aggregates local models in predictive space. The fact that the method is Bayesian means that it provides more accurate uncertainty estimates on predictions, and is therefore more robust in nature. The Bayesian perspective also provides the advantage of efficiency, since the technique operates in a single communication round. We performed experiments on various classification and regression datasets to show that the method performs competitively with other FL techniques, and that it outperforms them in more heterogeneous settings. We believe that taking a Bayesian perspective on the predictive space is a useful method for developing FL techniques due to its communication efficiency, and the fact that it side-steps the poor approximation issues of model space methods. 

\appendix

\bibliographystyle{named}
\bibliography{sources}

\end{document}